\documentclass[sigconf]{acmart}

\AtBeginDocument{%
  }

\acmSubmissionID{123-A56-BU3}

\copyrightyear{2026}
\acmYear{2026}
\setcopyright{cc}
\setcctype{by}
\acmConference[KDD '26]{Proceedings of the 32nd ACM SIGKDD Conference on Knowledge Discovery and Data Mining V.1}{August 09--13, 2026}{Jeju Island, Republic of Korea}
\acmBooktitle{Proceedings of the 32nd ACM SIGKDD Conference on Knowledge Discovery and Data Mining V.1 (KDD '26), August 09--13, 2026, Jeju Island, Republic of Korea}
\acmPrice{}
\acmDOI{10.1145/3770854.3783950}
\acmISBN{979-8-4007-2258-5/2026/08}

\author{Mingming Zhang}
\authornote{Both authors contributed equally to this research. Work done when Mingming Zhang and Na Li were interns at Taobao \& Tmall Group of Alibaba.}
\email{mingmingzhang@whu.edu.cn}
\affiliation{%
  \institution{Key Laboratory of Aerospace Information Security and Trusted Computing, Ministry of Education, School of Cyber Science and Engineering, Wuhan University \\ Taobao \& Tmall Group of Alibaba}
  \city{Wuhan}
  \country{China}
}

\author{Na Li}
\authornotemark[1]
\email{wannal@whu.edu.cn}
\affiliation{%
  \institution{Key Laboratory of Aerospace Information Security and Trusted Computing, Ministry of Education, School of Cyber Science and Engineering, Wuhan University \\ Taobao \& Tmall Group of Alibaba}
  \city{Wuhan}
  \country{China}
}

\author{Feiqing Zhuang}
\affiliation{%
  \institution{Taobao \& Tmall Group of Alibaba}
  \city{Hangzhou}
  \country{China}}
\email{zhuangfeiqing.zfq@alibaba-inc.com}

\author{Hongyang Zheng}
\affiliation{%
  \institution{Taobao \& Tmall Group of Alibaba}
  \city{Hangzhou}
  \country{China}}
\email{zhenghongyang.hy@alibaba-inc.com}

\author{Jiangbing Zhou}
\affiliation{%
  \institution{Taobao \& Tmall Group of Alibaba}
  \city{Hangzhou}
  \country{China}}
\email{zhoujiangbing.zjb@alibaba-inc.com}

\author{Wuyin Wang}
\affiliation{%
  \institution{Taobao \& Tmall Group of Alibaba}
  \city{Hangzhou}
  \country{China}}
\email{wangwuyin.wwy@alibaba-inc.com}

\author{Shengjie Sun}
\affiliation{%
  \institution{Taobao \& Tmall Group of Alibaba}
  \city{Hangzhou}
  \country{China}}
\email{shengjie.ssj@alibaba-inc.com  }

\author{Xiaowei Chen}
\affiliation{%
  \institution{Taobao \& Tmall Group of Alibaba}
  \city{Hangzhou}
  \country{China}}
\email{qisheng.cxw@alibaba-inc.com}

\author{Junxiong Zhu}
\affiliation{%
  \institution{Taobao \& Tmall Group of Alibaba}
  \city{Hangzhou}
  \country{China}}
\email{xike.zjx@alibaba-inc.com}

\author{Lixin Zou}
\email{zoulixin@whu.edu.cn}
\affiliation{%
  \institution{Key Laboratory of Aerospace Information Security and Trusted Computing, Ministry of Education, School of Cyber Science and Engineering, Wuhan University}
  \city{Wuhan}
  \country{China}
}

\author{Chenliang Li}
\authornote{Corresponding author.}
\email{cllee@whu.edu.cn}
\affiliation{%
  \institution{Key Laboratory of Aerospace Information Security and Trusted Computing, Ministry of Education, School of Cyber Science and Engineering, Wuhan University}
  \city{Wuhan}
  \country{China}
}

\renewcommand{\shortauthors}{Mingming Zhang et al.}

\usepackage{multirow}
\usepackage{booktabs}
\usepackage{enumitem}
\usepackage{graphicx}    % 插入图片需要
\usepackage{subcaption}  % 管理子图
\usepackage{caption}
\usepackage{xspace}
\usepackage{array}
\usepackage{balance}
\newcommand{\ProjectName}[1]{{\small\textsc{QGA}}} 

\newcommand{\eat}[1]{}
\newcommand{\paratitle}[1]{\vspace{1.5ex}\noindent \textbf{#1}}

\begin{document}

\title{Q-Regularized Generative Auto-Bidding: From Suboptimal Trajectories to Optimal Policies}

\begin{abstract}

With the rapid development of e-commerce, auto-bidding has become a key asset in optimizing advertising performance under diverse advertiser environments. The current approaches focus on reinforcement learning (RL) and generative models. These efforts imitate offline historical behaviors by utilizing a complex structure with expensive hyperparameter tuning. The suboptimal trajectories further exacerbate the difficulty of policy learning.

To address these challenges, we proposes \ProjectName{}, a novel \textbf{Q}-value regularized \textbf{G}enerative \textbf{A}uto-bidding method. In \ProjectName{}, we propose to plug a Q-value regularization with double Q-learning strategy into the Decision Transformer backbone. This design enables joint optimization of policy imitation and action-value maximization, allowing the learned bidding policy to both leverage experience from the dataset and alleviate the adverse impact of the suboptimal trajectories. Furthermore, to safely explore the policy space beyond the data distribution, we propose a Q-value guided dual-exploration mechanism, in which the DT model is conditioned on multiple return-to-go targets and locally perturbed actions. This entire exploration process is dynamically guided by the aforementioned Q-value module, which provides principled evaluation for each candidate action. Experiments on public benchmarks and simulation environments demonstrate that \ProjectName{} consistently achieves superior or highly competitive results compared to existing alternatives. Notably, in large-scale real-world A/B testing, \ProjectName{} achieves a 3.27\% increase in Ad GMV and a 2.49\% improvement in Ad ROI\footnote{https://github.com/M2C-Tech/QGA}.
\end{abstract}





\maketitle

\newcommand\kddavailabilityurl{https://doi.org/10.5281/zenodo.18030706}

\ifdefempty{\kddavailabilityurl}{}{
\begingroup\small\noindent\raggedright\textbf{KDD Availability Link:}\\
% please change the following context to include multiple artifacts if necessary.
The source code of this paper has been made publicly available at \url{\kddavailabilityurl}.
\endgroup
}

\section{Introduction}

\begin{figure}
    \centering
    \includegraphics[width=\linewidth]{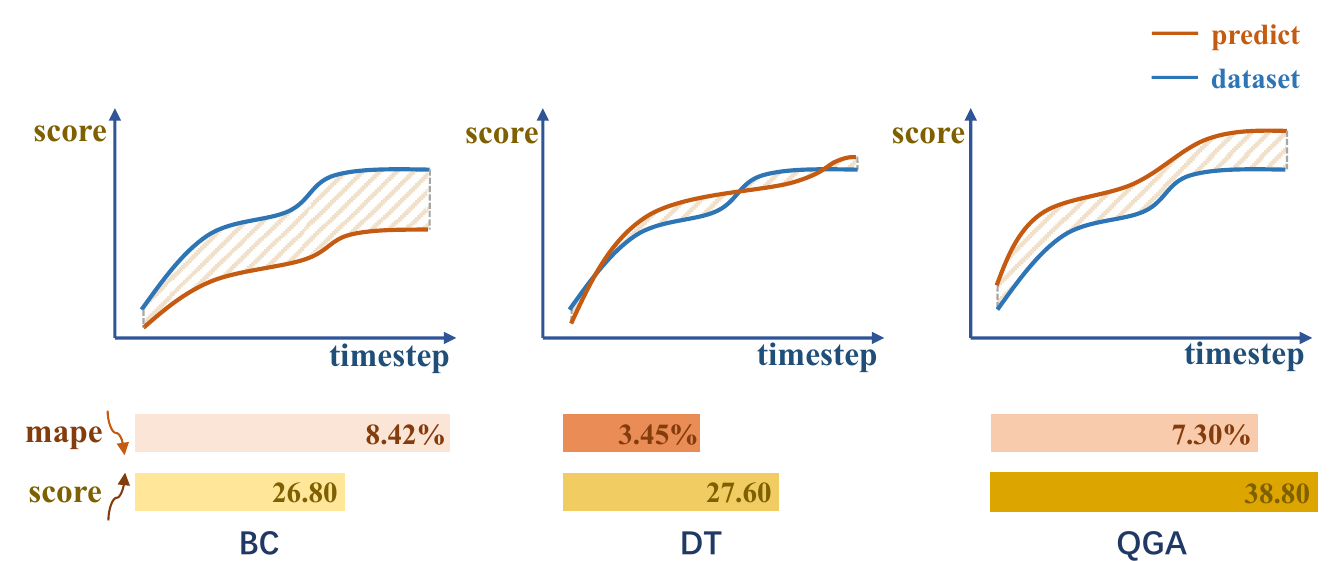}
    \caption{MAPE and performance scores of different models. Lower MAPE indicates stronger, while higher MAPE indicates poorer model imitation ability.}
    \label{fig:WMAPE and performance scores of different models.}
\end{figure}

Superior bidding strategies can bring better advertising performance and increase advertisers’ returns \cite{zhang2023personalized,li2024trajectory,liu2020effective,liu2020dynamic}. In modern advertising systems, the sheer volume of traffic, the rapid pace of market fluctuations, and the diverse preferences and demands of advertisers together create an extremely complex decision-making environment \cite{jha2024optimizing,aggarwal2024auto,korenkevych2024offline}. To meet these demands, bidding systems must be able to process a massive number of ad impressions in real time and provide reasonable bid prices under diverse requirements \cite{cai2025rtbagent,wang2023multi,li2022auto,yuan2022actor}. This renders traditional, manually-operated bidding methods inadequate for modern needs \cite{borissov2010automated,wen2022cooperative}.

Generative decision models, such as Decision Diffusion (DD)~\cite{guo2024generative} and Decision Transformer (DT)~\cite{chen2021decision}, effectively utilize historical behaviors for decision-making, have emerged as promising approaches for auto-bidding. However, these vanilla generative models highly rely on imitating offline behavioral policies over suboptimal behavioral trajectories. Figure \ref{fig:WMAPE and performance scores of different models.} shows the MAPE~\cite{de2016mean} and performance scores of different models on the offline dataset~\cite{su2024auctionnet}. The behavior cloning (BC) method~\cite{bain1995framework} with a shallow DNN has the highest MAPE and performs worst, while DT achieves the lowest MAPE but only moderate performance. Our method, with moderate MAPE, achieves the best performance instead, indicating that properly deviating from the behavior policy with exploration enhances results.

In recent years, generative bidding models equipped with exploration capabilities have demonstrated impressive performance~\cite{gao2025generative, li2025gas, guo2024generative, jiang2025optimal, li2025ebaret}. GAS~\cite{li2025gas} designed a post-training search mechanism based on Monte Carlo Tree Search~\cite{swiechowski2023monte, browne2012survey} and a value critic to expand the action space. GAVE~\cite{gao2025generative}, on the other hand, proposed a framework for value-guided optimization of DT itself. However, these methods also have notable limitations. GAS’s optimization only occurs during post-training search, and the process remains confined to a limited set of candidate actions. Furthermore, the backbone model (DT) essentially remains constrained by the quality of the behavioral policy, hindering true global optimization and strong generalization. GAVE involves multiple learning loss functions and complex action exploration mechanisms for policy optimization, requiring extensive hyperparameter tuning. This restricts its capacity to cope with complex and dynamic advertising environments~\cite{guo2024generative, he2021unified}, and its ability for out-of-distribution (OOD) policy exploration remains limited.

To address these obstacles, we propose a novel Q-regularized generative auto-bidding method with a more compact yet robust model architecture. Specifically, our approach introduces a Q-value regularization term into DT, enabling joint optimization of policy imitation and action value maximization. In this way, we can both leverage experience from the dataset and move beyond the constraints imposed by suboptimal trajectories, favoring better actions. Furthermore, to ensure safe OOD exploration, we design a dual policy exploration mechanism based on the Q-value module. By conditioning DT on a set of perturbed return-to-go (RTG) values, and perturbing multiple candidate actions jointly, we enable a dual search in both the RTG space and the local action neighborhood. The learned Q-value module can efficiently evaluate candidate actions and assist in selecting the optimal one, thereby facilitating the discovery of better bidding strategies in real-world advertising systems.

We conduct comprehensive empirical evaluations of the proposed method using both publicly available offline datasets and simulated advertising environments. Additionally, we validate its practical effectiveness through online A/B testing within large-scale production systems. The experimental results demonstrate strong and robust improvements over both traditional reinforcement learning (RL) methods and state-of-the-art generative approaches, particularly on the challenging AuctionNet-Sparse dataset, as measured by key business metrics.

In summary, our principal contributions are as follows:
\begin{itemize}
\item We propose a novel Q-regularized generative auto-bidding method that simultaneously enables effective policy imitation and adaptive policy improvement, providing strong support for auto-bidding under diverse advertiser demands. 
\item We further propose a dual policy exploration mechanism for better OOD exploration. This cross search driven by the multi-RTG conditioning with action perturbation can help us go beyond the suboptimal datasets for superior bidding strategies.
\item We provide comprehensive experimental evaluation using public offline datasets, simulation environments and large-scale real-world advertising platforms. The results demonstrate significant and robust improvements over both state-of-the-art generative and mainstream bidding algorithms, particularly on the challenging AuctionNet-Sparse dataset, across key business metrics.
\end{itemize}

\section{Related Works}

\textbf{Advertisement Auto-bidding.}
The development of automated bidding techniques, a cornerstone of modern advertising auctions, can be divided into four stages. Early methods used Proportional-Integral-Derivative (PID) controllers and their variants \cite{chen2011real, zhang2016feedback, yang2019bid}, optimizing budget consumption as a proxy for performance, but proved inadequate for complex scenarios. Subsequently, rule-based algorithms \cite{yu2017online, yu2020low} emerged, forecasting auction traffic and directly optimizing objectives through parameter selection. The third stage introduced reinforcement learning (RL) approaches, ranging from classical RL \cite{zhao2018deep, cai2017real, wu2018budget} to offline \cite{he2021unified} and online RL methods \cite{mou2022sustainable}. Although these RL methods have achieved good results, they do not perform well in modeling long-term trajectories under complex scenarios. The recent automatic bidding solutions are based on generative decision models. DiffBid is a novel generative advertising bidding model based on the Decision Diffusion~\cite{guo2024generative}. Subsequently, several models leveraging the Decision Transformer, such as GAVE \cite{gao2025generative} and GAS \cite{li2025gas}, have been introduced. However, GAVE suffers from high structural complexity and its performance heavily relies on extensive hyperparameter tuning, while GAS depends on post-training search over the limited candidate actions. 
\eat{
 In this work, during training, we use a Q-value regularization term to enable DT to learn high-quality actions in the dataset rather than simply imitate the behavioral policy. During inference, we design a mechanism that leverages the Q-value module to explore better OOD policies, thereby improving bidding performance.
}

\textbf{Reinforcement Learning and Generative Decision Making.}
RL has achieved significant success in sequential decision-making tasks \cite{mnih2015human, schulman2015trust, levine2020offline}. Offline RL enables policy learning from static datasets, which is valuable in high-risk or high-cost domains. To address distribution shift and value overestimation issues in offline RL, approaches like BCQ~\cite{fujimoto2019off}, CQL~\cite{kumar2020conservative}, and IQL~\cite{kostrikov2021offline} have been developed. Nevertheless, traditional methods are limited by their reliance on Markov Decision Processes (MDP), restricting their ability to model long-term dependencies.

Recently, generative decision-making approaches have emerged by framing RL as a sequence modeling problem. The Decision Transformer (DT)~\cite{chen2021decision} utilizes transformer architectures to capture temporal and historical patterns, achieving state-of-the-art results in offline RL. Further extensions, such as the Conditional Decision Transformer (CDT)~\cite{liu2023constrained}, enable constraint adaptation and improved generalization. QT \cite{hu2024q} leverages a Q-value guidance mechanism to enhance the ability of DT to learn better actions. Inspired by this line, our work exploit a Q-value regularization based DT framework and design a novel dual policy exploration mechanism for exploring OOD actions.

\section{Preliminary}
\subsection{Problem Formulation: Auto-Bidding}

\subsubsection{Auction Environment}

We consider an auction environment in which a sequence of $I$ impression opportunities arrives sequentially at discrete time steps $i = 1, \ldots, I$. At each time step $i$, advertisers participate by submitting bids denoted as $\{b_i\}$. An advertiser wins the $i$-th impression if their bid $b_i$ exceeds the highest bid submitted by other competitors, denoted as $b_i^-$. Under the generalized second-price (GSP) auction mechanism\cite{edelman2007internet}, the winning advertiser pays a cost $c_i$, which is equal to the second-highest bid. For each impression won, the advertiser receives a private value $v_i$, often determined by metrics such as the predicted conversion probability or click-through rate.

\subsubsection{Optimization Objective and Constraints}

During an advertising period, the advertiser's primary objective is to maximize the total value obtained from winning impression opportunities, subject to both budget and performance constraints. Let $x_i \in \{0, 1\}$ indicate whether impression $i$ is won ($x_i = 1$) or not ($x_i = 0$). The optimization problem can be formalized as follows:
\begin{equation}
\begin{aligned}
    \max_{x_i} \quad & \sum_{i=1}^{I} x_i v_i \\
    \text{s.t.} \quad 
        & \sum_{i=1}^{I} x_i c_i \leq B \\
        & \frac{\sum_{i=1}^{I} x_i c_i}{\sum_{i=1}^{I} x_i v_i} \leq C \\
        & x_i \in \{0, 1\}, \quad \forall i
\end{aligned}
\end{equation}
where $v_i$ denotes the value of impression $i$, $c_i$ is the corresponding cost, $B \in \mathbb{R}^+$ represents the total budget, and $C \in \mathbb{R}^+$ specifies the maximum allowable cost per acquisition (CPA). 

This formulation ensures that the advertiser neither exceeds the allocated budget nor surpasses the predefined upper limit for average cost per acquisition. Such a multi-constraint framework is crucial for balancing cost efficiency and advertising effectiveness.

\eat{
\subsubsection{Analytical Bidding Solution}

\begin{equation}
    b_i^* = \lambda_0^* v_i - \sum_{j} \lambda_j^* \left(q_{ij}(1 - \mathbb{1}_{CR_j}) - k_j p_{ij}\right)
\end{equation}
where $\lambda_0^*$ and $\lambda_j^*$ are the optimal Lagrangian multipliers corresponding to the constraints, $v_i$ is the value estimate for impression $i$, and $q_{ij}$, $p_{ij}$, and $k_j$ represent parameters related to various performance indicators and campaign constraints. The indicator function $\mathbb{1}_{CR_j}$ identifies whether constraint $j$ is cost-related.

}

\subsubsection{Analytical Bidding Solution}

Previous research~\cite{he2021unified} reformulated the auto-bidding problem as a linear programming (LP) problem, allowing for an analytical solution to the optimal bidding policy. In the case of a single CPA constraint, the optimal bid simplifies to:
\begin{equation}
    b_i^* = \lambda^* v_i, \quad \text{where} \quad \lambda^* = \lambda_0^* + \lambda_1^* C
\end{equation}
This reformulation reduces the control problem to dynamically learning or adjusting the unified bidding parameter $\lambda^*$ to optimize performance while satisfying all constraints~\cite{he2021unified,su2024auctionnet}.

\begin{figure*}[htbp]
    \centering
    \includegraphics[width=0.78\textwidth]{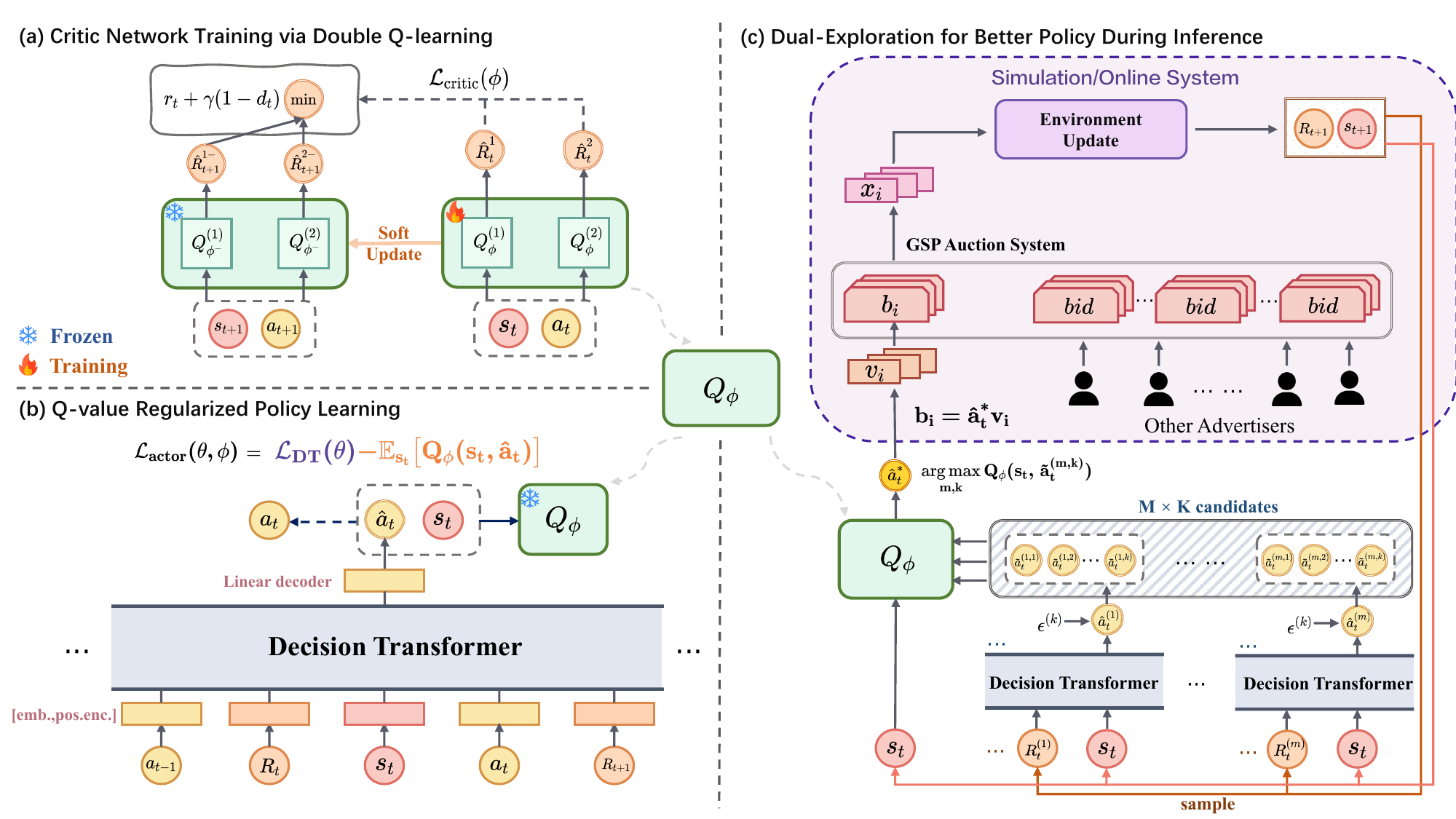} 
    \caption{Overall structure of \ProjectName{}.}
    \label{fig:overall_structure}
\end{figure*}

\subsection{Sequential Decision Modeling}

In the rapidly evolving field of online advertising, bidding policies must dynamically adapt to optimize campaign performance. Traditional static or myopic approaches often fail to account for auction volatility, fluctuating budget constraints, and time-sensitive objectives. Drawing inspiration from generative bidding frameworks, we thus conceptualize auto-bidding as a \emph{sequential decision-making} problem, wherein each decision influences outcomes throughout the entire campaign horizon.

\paratitle{Formalization as a Markov Decision Process.}
We model auto-bidding as a sequential decision process in which an agent interacts with the advertising environment $E$ at discrete time steps. Specifically, at each time step $t$: 1)~The agent observes the current \textbf{state} $s_t \in \mathcal{S}$, which encapsulates relevant contextual information; 2)~Based on $s_t$, the agent selects an \textbf{action} $a_t \in \mathcal{A}$, representing the bid amount; 3)~The environment transitions according to underlying, potentially unknown, dynamics $\mathcal{T}$. In the Markov Decision Process (MDP) framework, the transition function $\mathcal{T}: s_t \times a_t \rightarrow s_{t+1}$ indicates that the next state $s_{t+1}$ is determined solely by the current state and action. On contrary, the transition for auto-bidding may depend on the full trajectory $\tau$; 4)~The agent receives a \textbf{reward} $r_t$, quantifying the value obtained at time $t$. This process is repeated until the end of the bidding period (e.g., a day), with the primary objective of maximizing the cumulative reward.

\paratitle{Modeling Auto-Bidding.}
According to the above setting, we further refine the auto-bidding problem as follows:
\begin{itemize}
    \item \textbf{State $s_t$:} A feature vector representing the contextual state at time $t$, including remaining campaign duration, available budget, and historical bidding statistics.
    \item \textbf{Action $a_t$:} The bidding parameter selected at time $t$, denoted as $a_t = \lambda_t$, consistent with the optimal bidding policy (\textit{cf.} Eq.~(3)).
    \item \textbf{Reward $r_t$:} Given $N_t$ impressions delivered between $t$ and $t+1$, the reward is defined as $r_t = \sum_{n=1}^{N_t} x_{nt} v_{nt}$, representing the total value accrued during that interval.
    \item \textbf{Return-to-Go $R_t$:} The cumulative anticipated reward from time $t$ onward, defined as $R_t = \sum_{t' = t}^{T} r_{t'}$, where $T$ denotes the campaign horizon.
\end{itemize}

\section{Method}
\eat{
The core challenge in advertising bidding is to learn an improved policy from a large number of suboptimal bidding trajectories. We propose \ProjectName{}, which utilizes the trajectory modeling capability of DT and leverages behavior data generated by existing policies as a foundation. Additionally, we introduce a Q-value regularization module to guide DT towards superior actions within the dataset, and design a Q-value-based policy exploration mechanism to discover policies not present in the existing data. Consequently, \ProjectName{} can derive policies that outperform the current behavior policy. 
}

In this section, we first present a detailed explanation of constructing DT as the fundamental backbone of the bidding model. Then, we describe the incorporation of Q-value regularization into policy learning. Finally, we discuss how to leverage the Q-value regularized model on the existing dataset to explore policies that extend beyond the current behavior policy.

\subsection{DT for Bidding Policy Learning}
Here, we employ DT \cite{chen2021decision} as the backbone model for learning sequential bidding policies. DT is a trajectory-based model that formulates sequential decision-making as a conditional sequence modeling problem, leveraging the powerful representation learning capabilities of Transformer architectures.

\subsubsection{Trajectory Formulation}
To adapt DT for the auto-bidding task, we first represent the agent's experience as a collection of trajectories. Each trajectory $\tau$ consists of a sequence of $(R_t, s_t, a_t)$ tuples.
During training, the DT model learns to predict the optimal next action $a_t$ conditioned on the desired future return $R_t$ and the observed state $s_t$. This sequence modeling approach enables DT to learn the bidding policy of the existing trajectory.

\subsubsection{Sequence Modeling with Decision Transformers}  
The Decision Transformer uses causal Transformer blocks to process the sequential input tokens. At each time step $t$, the model receives embeddings of $(R_t, s_t, a_t)$. In practice, the input to DT consists of the advertising bidding trajectory from the most recent $K$ steps.
\begin{equation}
    (R_{t-K+1}, s_{t-K+1}, a_{t-K+1}, R_{t-K+2}, s_{t-K+2}, a_{t-K+2}, \ldots, R_{t}, s_{t})
\end{equation}
These embeddings are processed by the Decision Transformer to produce contextualized representations, from which the model predicts the next action $a_{t+1}$ based on the history up to time $t$ and the target return-to-go $R_{t+1}$.

\subsubsection{Training Objective}
The model is trained via a supervised learning objective: given a dataset of historical bidding trajectories, the Decision Transformer is optimized to minimize the mean squared error (MSE) between the predicted action $\hat{a}_t$ and the ground-truth action $a_t$ taken in the observed trajectory:

\begin{align}
    \mathcal{L}_{DT} 
    &=\mathbb{E}_{\tau \sim \mathcal{D}} \;
        \mathbb{E}_{t_0 \sim \mathcal{U}[1, T-K+1]}  \notag \\
    \frac{1}{K} & \sum_{k=0}^{K-1} 
    \mathbb{E}\Big[ 
        (\hat{a}_{t_0+k} - a_{t_0+k})^2 \Big|~ (R_{t_0 : t_0+K},\, s_{t_0 : t_0+K},\, a_{t_0 : t_0+K-1}) 
    \Big]
\end{align}

By conditioning on various values of $R_t$ during training, the Decision Transformer can generate the action with respect to different levels of campaign performance, which supports controllable trade-offs between value maximization and constraint satisfaction.

\subsection{Q-value Regularized Policy Learning}
Although the Decision Transformer framework enables behavioral cloning from offline trajectories, its performance is fundamentally limited by the quality of the behavior policy in the dataset. The standard Decision Transformer only learns to imitate observed actions, lacking any explicit mechanism to favor actions that could lead to higher returns. To address this limitation, inspired by \cite{hu2024q}, we introduce a modification that allows the policy to utilize value estimates from an auxiliary critic network. This enhancement enables the generation of trajectories with higher expected returns than imitation alone.

\subsubsection{Actor Loss with Q-value Regularization}

Let $s_t$ and $a_t$ denote the state and action at time $t$, and let $\hat{a}_t$ represent the action predicted by the Decision Transformer given the current historical context. To encourage the Decision Transformer to predict high-return actions, we introduce a critic network $Q_\phi(s_t, a_t)$, parameterized by $\phi$, which estimates the expected return for each state-action pair. The regularization term for the actor loss is defined as:
\begin{align}
    \mathcal{L}_Q(\phi) &= -\mathbb{E}_{s_t}\big[ Q_\phi(s_t, \hat{a}_t) \big]
\end{align}
Combining this with the original Decision Transformer loss, we define the overall actor loss as:
\begin{equation}
    \mathcal{L}_{\text{actor}}(\theta, \phi) = \mathcal{L}_{\text{DT}}(\theta) + \alpha \cdot \mathcal{L}_Q(\phi)
\end{equation}
where $\theta$ are the parameters of the Decision Transformer, and $\alpha > 0$ is a regularization coefficient. In this way, the Decision Transformer is trained both to imitate actions from the dataset and to maximize the predicted return according to the learned critic.

\subsubsection{Critic Training} 
The critic network serves as an estimator for the action-value function $Q_{\phi}(s_t, a_t)$, which is essential for providing value-based feedback to the actor. In accordance with the principles of double Q-learning\cite{hasselt2010double,van2016deep}, we maintain two distinct Q-networks, $Q_{\phi}^{(1)}$ and $Q_{\phi}^{(2)}$, with the same parameters $\phi$. During training, for each transition in the batch, we obtain $(s_t, a_t, r_t, s_{t+1}, d_t)$, where $r_t$ is the observed reward and $d_t$ is a binary indicator for episode termination.

To ensure stable value estimation, the target Q-value is calculated using a separate target network, whose parameters, $\phi^{-}$, are an exponential moving average of the main critic’s parameters\cite{lillicrap2015continuous, fujimoto2018addressing}. At each step, we compute the predicted next action $\hat a_{t+1}$ as the actor's output for state $s_{t+1}$ in the sequence. The Bellman target \cite{bellman1954theory} for each transition is then given by:
\begin{equation}
    y_t = r_t + \gamma (1-d_t) \cdot \min_{i \in \{1,2\}} Q_{\phi^-}^{(i)}(s_{t+1}, a_{t+1})
\end{equation}
where $\gamma$ denotes the discount factor. By taking the minimum of the two target Q-values, we mitigate positive bias and obtain more conservative and reliable targets.

The critic loss aggregates the squared temporal-difference (TD) errors for both Q-heads over the batch:
\begin{equation}
    \mathcal{L}_{\text{critic}}(\phi) = \frac{1}{N} \sum_{j=1}^N \left[
        \left(Q_{\phi}^{(1)}(s_{t_j}, a_{t_j}) - y_{t_j}\right)^2 + \left(Q_{\phi}^{(2)}(s_{t_j}, a_{t_j}) - y_{t_j}\right)^2
    \right]
\end{equation}
where $N$ denotes the number of valid non-terminal positions within the batch.

After computing gradients of $\mathcal{L}_{\text{critic}}$ with respect to $\phi$, we update the critic network parameters via gradient descent. To promote temporal consistency and smoothness of the targets, the target network parameters are updated at each step via exponential moving averaging:
\begin{equation}
    \phi^- \leftarrow (1 - \tau) \phi^- + \tau \phi
\end{equation}
where $\tau$ is a small positive value.

The critic training method enables the Q-network to accurately estimate values based on  offline trajectories, thereby providing essential support to \ProjectName{}. During training, Q-value regularization by the critic not only guides the actor to imitate the dataset policy but also encourages the maximization of action values, thereby integrating value-based objectives into the learning process. Furthermore, within the policy exploration mechanism, the critic functions as an offline value estimator to evaluate and rank diverse candidate actions.

% \begin{theorem}
% Let $\pi$ denote the policy learned by standard DT, and let $\pi^*$ denote the policy learned by Q-value regularized DT. Under mild assumptions on the Q-function approximator and for an appropriately chosen regularization coefficient $\alpha > 0$, the expected return of $\pi^*$ is at least no less than that of $\pi$, i.e.,
% \begin{equation}
%     J(\pi^*) \geq J(\pi)
% \end{equation}
% where $J(\pi)$ denotes the expected return when following policy~$\pi$.
% \end{theorem}

\subsection{Dual-Exploration for Better Policy}
Although Q-value regularization encourages the Decision Transformer to select actions with higher estimated value within the offline dataset, the resulting policy is still fundamentally constrained by fragments of the behavior trajectories\cite{hepburn2022model, li2024diffstitch}. No matter training or inference, the actions generated by the Decision Transformer typically stay close to those in the historical trajectories or are stitching of fragments from these trajectories. This constraint limits its capacity to explore and identify novel or potentially superior bidding policies.

\subsubsection{Q-value based Policy Improvement}
To further alleviate the limitations of behavior cloning and Q-value regularization, we introduce a policy improvement strategy during inference by combining \textbf{multi-return-to-go (multi-rtg) conditioning} with \textbf{action perturbation}. This approach enables broader and safer off-policy exploration, substantially enhancing the agent's ability to identify high-value actions even outside the empirical support of the offline dataset.

Concretely, for each decision step $t$, our method proceeds as follows:
\begin{enumerate}
    \item \textbf{Multi-RTG Sampling:} We sample a set of $M$ different candidate return-to-go values $R_t^{(1)}, \ldots, R_t^{(M)}$, usually centered around the current estimated return. Each return-to-go represents a hypothetical future performance target for the Decision Transformer model.
    \item \textbf{Action Generation:} For each candidate $R_t^{(m)}$, the Decision Transformer predicts an action $\hat{a}_t^{(m)}$ conditioned on $R_t^{(m)}$ and the episode history.
    \item \textbf{Action Perturbation:} To further diversify the action space, we add small random noise to each predicted action to obtain $K$ perturbed actions, i.e., $\tilde{a}_t^{(m, k)} = \hat{a}_t^{(m)} + \epsilon^{(k)}$, where $\epsilon^{(k)}$ is drawn from Gaussian noise distribution.
    \item \textbf{Q-value Evaluation and Selection:} All candidates (across different $R_t^{(m)}$ and perturbations $k$) are evaluated by the learned critic network to obtain Q-value estimates $q^{(m, k)} = Q_\phi(s_t, \tilde{a}_t^{(m, k)})$. The final deployed action is the one with the highest estimated Q-value:
    \begin{equation}
        \hat{a}_t^* = \mathop{\arg\max}_{m, k} Q_\phi(s_t,\, \tilde{a}_t^{(m, k)}).
    \end{equation}
\end{enumerate}

This process effectively constructs a two-dimensional search over possible future objectives (through multi-rtg conditioning) and local action variations (via perturbation), all guided by the offline-trained value function. It enables a significantly richer and more adaptive exploration of the action space compared to naive imitation or single-rtg Q-regularized methods. Our policy improvement approach provides a principled and practical strategy for substantially increasing the likelihood of discovering novel actions with superior long-term value in auto-bidding.

\begin{table*}[htbp]
    \setlength{\tabcolsep}{8pt} % 在这里调整列间距
    \centering
    \caption{Offline evaluation on the AuctionNet and AuctionNet-Sparse datasets. Bold indicates the best results, underline denotes the second-best results, and "*" marks results obtained directly from the original paper.}
    \begin{tabular}{c|c|cccccccccc}
    
\hline\hline
        Dataset & Budget & IQL & BC & CQL & TD3-BC & DT & DiffBid & $\text{GAS}^*$ & $\text{GAVE}^*$  & \ProjectName{}\\
        \hline
        \multirow{5}{*}{AuctionNet~\cite{su2024auctionnet}} 
            & 50\%  & 165 & 169  & 77  & 167  & 170 & 72 & 193 & \underline{201} &  \textbf{206}\\
            & 75\%  & 234 & 236  & 116 & 239  & 232 & 129 & \underline{287} & \textbf{296} &  282\\
            & 100\% & 293 & 285  & 158 & 288  & 298 & 169 & \underline{359} & \textbf{376} &  353\\
            & 125\% & 331 & 342  & 194 & 332  & 340 & 221 & \underline{409} & \textbf{421} &  \underline{409}\\
            & 150\% & 378 & 375  & 231 & 376  & 379 & 280 & 461 & \textbf{467} &  \underline{463}\\
        \hline
        \multirow{5}{*}{AuctionNet-Sparse~\cite{su2024auctionnet}} 
            & 50\%  & 17.9 & 15.0 & 16.1 & 15.0 & 18.4 & 10.7 & \underline{18.4} & \textbf{19.6} & \textbf{19.6}\\
            & 75\%  & 26.9 & 20.3 & 22.4 & 22.7 & 24.9 & 22.2 & 27.5 & \underline{28.3} &  \textbf{29.0}\\
            & 100\% & 30.9 & 26.8 & 27.9 & 26.4 & 27.6 & 24.6 & 36.1 & \underline{37.2} &  \textbf{38.8}\\
            & 125\% & 32.0 & 31.6 & 32.1 & 31.4 & 35.6 & 31.8 & 40.0 & \underline{42.7} &  \textbf{45.3}\\
            & 150\% & 37.8 & 36.6 & 37.6 & 38.0 & 39.4 & 36.5 & 46.5 & \underline{47.4} &  \textbf{50.1}\\
        \hline\hline
    \end{tabular}
    \label{tab:Offline Test on AuctionNet datasets}
\end{table*}

\section{Experiments}
In this section, we present extensive evaluations of the proposed \ProjectName{}, including both offline testing and testing in a simulation environment. Comprehensive ablation studies and parameter experiments are also carried out to address the following questions:

\begin{itemize}[leftmargin=10pt]
    \item RQ1: How does \ProjectName{} perform compared to other traditional and generative baseline methods?  
    \item RQ2: Can \ProjectName{} adapt to more complex and realistic environments?
    \item RQ3: How do the proposed core components contribute to the performance of \ProjectName{}?  
    \item RQ4: How do different parameter settings affect the performance?
\end{itemize}
 
\subsection{Datasets}

To ensure fairness and reproducibility, we evaluated on AuctionNet~\cite{su2024auctionnet}, a large-scale, publicly available benchmark released by Alimama and widely used for advertising auto-bidding research.

Specifically, we used two datasets: \emph{AuctionNet} and \emph{AuctionNet-Sparse}, corresponding to dense and sparse conversion scenarios, respectively. Each dataset contains about 480,000 bidding trajectories, with 48 time steps in each trajectory. In addition, we utilized the AuctionNet simulation environment for realistic evaluation. More dataset details are provided in Appendix~\ref{app:Offline Dataset}.

\subsection{Evaluation Settings}
\subsubsection{Metrics.} We evaluated model effectiveness using the advertising bidding score, which reflects the agent’s ability to maximize conversions while adhering to the advertiser’s CPA constraint $C$. If the actual CPA exceeds $C$, a penalty is imposed. The score is defined as:
\begin{equation}
\label{equ:Score}
    Score = \mathbb{P}(CPA;C) \cdot \sum_{i} x_{i} \cdot V_{i}
\end{equation}
the penalty function for exceeding the CPA constraint is given by:
\begin{equation}
\label{equ:P(CPA;C)}
    \mathbb{P}(CPA;C) = \min \left\{ \left( \frac{C}{CPA} \right)^{\beta}, \ 1 \right\}
\end{equation}
with $\beta > 0$ (typically,$\beta =2$). The penalty is only enforced when $CPA > C$.

\subsubsection{Evaluation Protocol} 
We followed the methodology of AuctionNet \cite{su2024auctionnet}. During offline evaluation, a bid is considered successful if it exceeds the historical minimum winning price recorded in the test dataset. To comprehensively evaluate the performance of one specific model across different advertisers, we sequentially used it to act as the agent for each advertiser while the remaining advertisers served as competitors. The agent's final score is computed using Equation~\ref{equ:Score} at the end of the delivery period.

AuctionNet also provides a standardized simulated online environment, where ad delivery consists of 48 decision time steps. At each step, multiple advertisers bid for numerous impression opportunities. Further details on the simulation environment are presented in Appendix~\ref{app:Simulation Environment}.

\subsubsection{Baselines}
We evaluate \ProjectName{} on the auto-bidding task against several representative baselines covering offline reinforcement learning and generative modeling paradigms. For offline reinforcement learning, we include \textbf{BC \cite{bain1995framework}}, \textbf{IQL \cite{kostrikov2021offline}}, \textbf{CQL \cite{kumar2020conservative}} and \textbf{TD3-BC \cite{fujimoto2021minimalist}}. For generative methods, we considere \textbf{DiffBid \cite{guo2024generative}} , which is based on Decision Diffusion, as well as \textbf{DT~\cite{chen2021decision}} and its variants, \textbf{GAS~\cite{li2025gas}} and \textbf{GAVE~\cite{gao2025generative}}.

\subsubsection{Deployment Settings}
All traditional reinforcement learning baselines~\cite{kostrikov2021offline, kumar2020conservative, fujimoto2021minimalist} were implemented using the architectures and parameter settings specified in the AuctionNet benchmark~\cite{su2024auctionnet}. DiffBid~\cite{guo2024generative} was implemented according to the original paper's details and settings. For GAVE~\cite{gao2025generative}, we reuse the original paper's performance as we couldn't reproduce its results.

\subsection{RQ1: Evaluation Results of Offline Testing}

Table~\ref{tab:Offline Test on AuctionNet datasets} reports offline test results for \ProjectName{} and multiple baselines on both AuctionNet and AuctionNet-Sparse datasets under different budget constraints. Our key findings are as follows:

\textbf{Consistently Superior Performance:} \ProjectName{} consistently achieves either the best or highly competitive results across all budget levels on both datasets. Notably, on the more challenging AuctionNet-Sparse dataset, \ProjectName{} exhibits pronounced superiority, particularly as the budget increases. The Sparse dataset, with its sparse conversion rates, more closely resembles real-world scenarios. These results demonstrate \ProjectName{}’s strong adaptability and generalization in offline bidding environments—especially under sparse conversion circumstances—thus providing better alignment with practical, real-world settings.

\textbf{Comparison with DT-based Methods:} While Decision Transformer and its variants (e.g., GAS, GAVE) outperform traditional RL baselines due to their advanced sequence modeling abilities, QGA consistently matches or exceeds the performance of these methods, particularly at higher budget levels. The performance improvements observed on the AuctionNet dataset are less substantial than those on the AuctionNet-Sparse dataset, which may be attributed to the much higher conversion rate in AuctionNet (Appendix \ref{app:Offline Dataset}), thereby diminishing the relative impact of policy optimization.

\textbf{Imitation behavior cannot produce better policies:} BC and DT, which rely solely on imitation, underperform compared to methods with exploration (e.g., IQL, GAS, GAVE). This indicates that imitation learning from suboptimal data is insufficient, and exploration mechanisms are crucial for better performance.

\textbf{Suboptimal Performance of DiffBid:} DiffBid underperforms on both datasets, likely due to sparse rewards and dynamic, long-horizon environments. Limited reward signals impede learning, and changing dynamics undermine effective reverse process inference, resulting in poor trajectory and action quality.

\subsection{RQ2: Evaluation Results in Simulation Environment}

To evaluate the bidding capabilities of \ProjectName{} in complex and realistic auction scenarios, we conducted experiments in simulated environments. As demonstrated by the results in Table~\ref{tab:methods_metrics_sim_env}, \ProjectName{} achieves a score of 8113 in the simulation environment, substantially outperforming all baseline methods, including IQL (6534), CQL (7138), and GAS (7454). This significant performance improvement indicates that \ProjectName{} is highly effective at adapting to the complexities of the simulation environment. These results provide strong empirical evidence that \ProjectName{} can generalize well and maintain high performance in complex, dynamic simulated settings.

\subsection{RQ3: Ablation Study}

\begin{figure}
    \centering
    \includegraphics[width=0.9\linewidth]{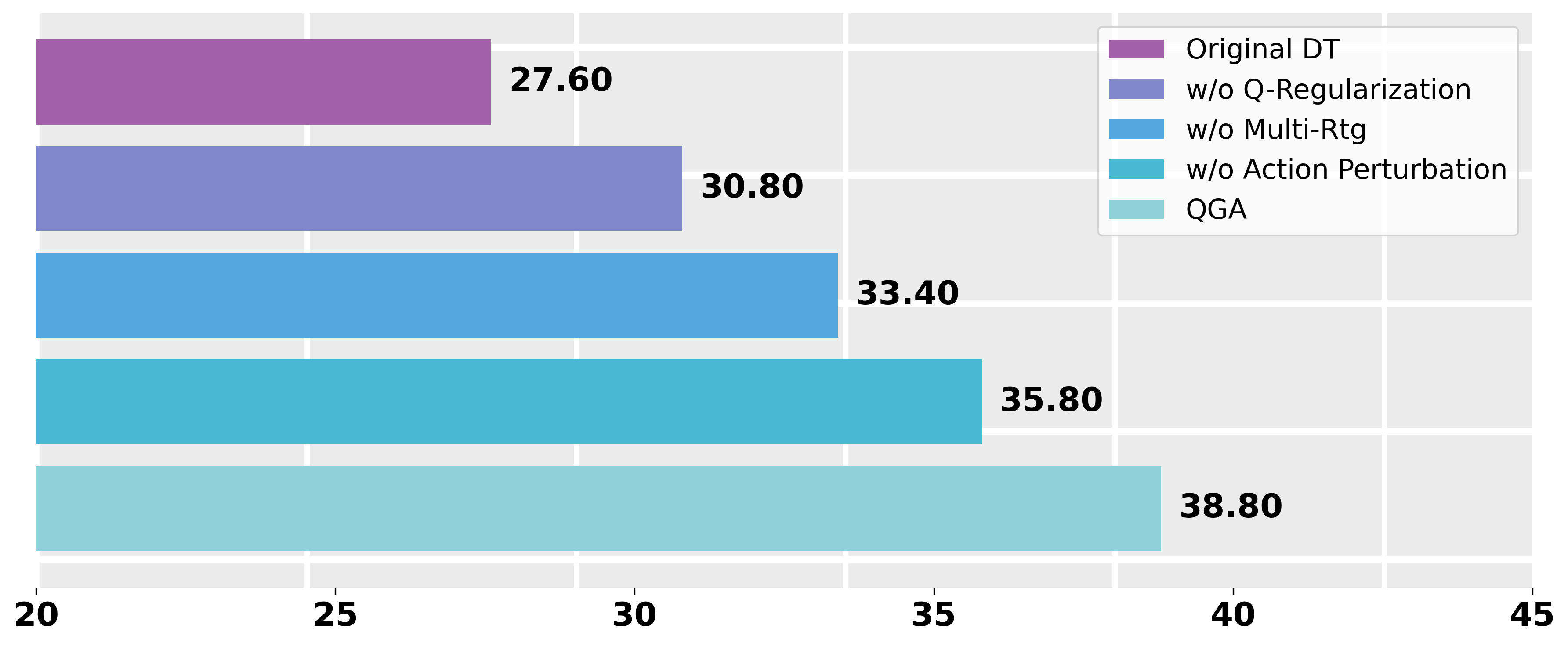}
    \caption{Ablation study}
    \label{fig:ablation_sim_env}
\end{figure}

Figure~\ref{fig:ablation_sim_env} presents the results of the ablation study. \ProjectName{}, with the full configuration, achieves the highest score of 38.8. When either the Action Perturbation or Multi-RTG module is removed, the scores decrease to 35.8 and 33.4, respectively. Removing Q-Regularization leads to the largest drop in performance, resulting in a score of 30.8. In contrast, the original baseline DT model attains only 27.6. These findings demonstrate that each proposed module independently enhances model performance, and their combination yields complementary effects that further improve overall results. This ablation analysis confirms the necessity and synergy of both the Q-Regularization and Dual-Exploration module in improving the effectiveness of the model.

\begin{table}[htbp]
    \centering
    \caption{Performance in the Simulation Environment}

    \begin{tabular}{
        >{\centering\arraybackslash}m{2cm} 
        >{\centering\arraybackslash}m{2cm} 
        >{\centering\arraybackslash}m{2cm}
    }
        \toprule
        %\rowcolor{gray!15}
        Dataset & Method & Score \\
        \midrule
        \multirow{9}{*}{Simulation Env.} 
            & IQL      & 6534 \\
            & BC       & 6366 \\
            & CQL      & 7138 \\
            & TD3-BC   & 7008 \\
            & DT       & 6920 \\
            & DiffBid  & 6248 \\
            & GAS      & 7454 \\
            & \textbf{OURS}     & \textbf{8113} \\
        \bottomrule
    \end{tabular}
    \label{tab:methods_metrics_sim_env}
\end{table}

\eat{
\begin{table}[htbp]
    \centering
    \caption{Performance in the Simulation Environment}
    \begin{tabular}{c|c|c}
        \hline\hline
        Dataset & Method & Score \\
        \hline\hline
        \multirow{10}{*}{Simulation Env.}
                & IQL    & 6534 \\
                & BC     & 6366 \\
                & CQL    & 7138 \\
                & TD3-BC & 7008 \\
                & DT     & 6920 \\
                & DiffBid     & 6248 \\
                & GAVE   & -    \\
                & GAS    & 7454 \\
                & OURS   & \textbf{8113} \\
        \hline\hline
    \end{tabular}
    \label{tab:methods_metrics_sim_env}
\end{table}
}

\subsection{RQ4: Hyper-parameter Study}

We conducted a comprehensive hyper-parameter study to investigate the effects of the regularization coefficient $\alpha$ in the actor loss, as well as the number of return-to-go samples ($M$) and action samples ($N$) utilized in the dual-exploration module.

\begin{figure}[!t]
    \centering
    % 第一个子图
    \begin{subfigure}[b]{0.44\linewidth}
        \centering
        \includegraphics[width=\linewidth]{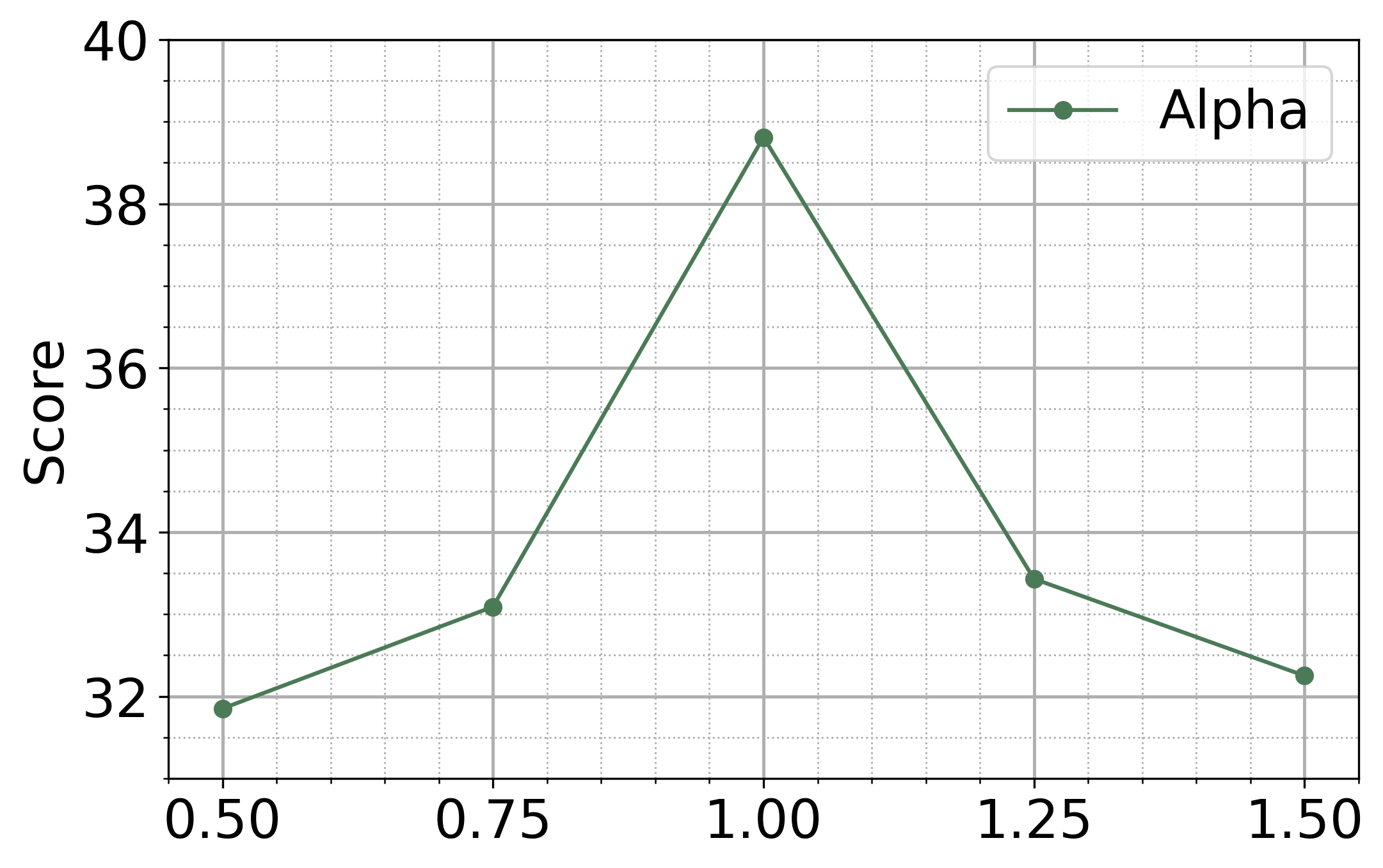} % 换成你的图片文件
        \caption{Regularization Coefficient $\alpha$}
        \label{fig:score_vs_alpha}
    \end{subfigure}
    \hspace{6pt} % 或尝试更小/更大的数值
    % 第二个子图
    \begin{subfigure}[b]{0.45\linewidth}
        \centering
        \includegraphics[width=\linewidth]{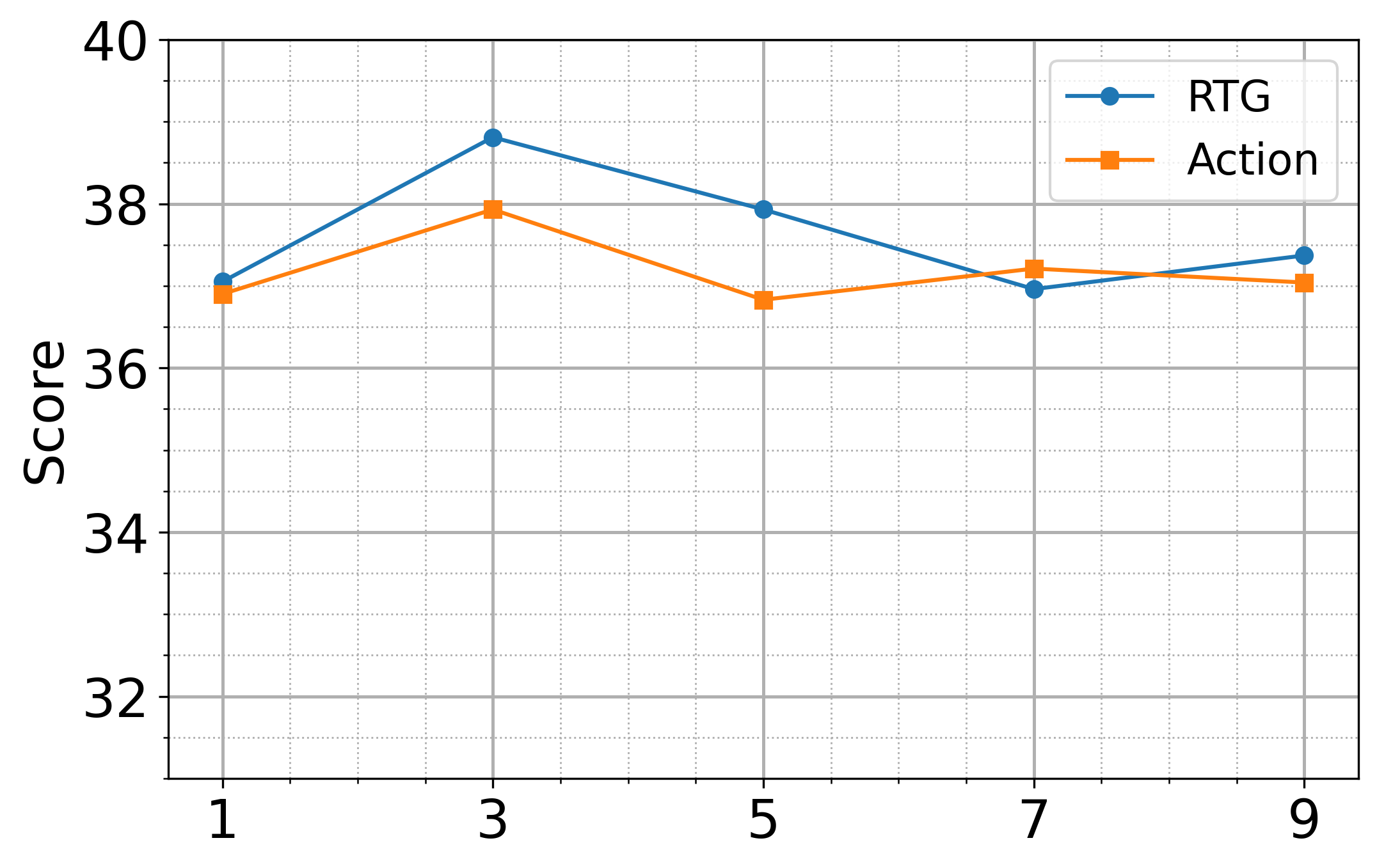} % 换成你的图片文件
        \caption{The Number of RTG and Actions}
        \label{fig:score_mn}
    \end{subfigure}
    \caption{Hyper-parameter Study}
    \label{fig:Hyper-parameter Study}
\end{figure}

\textbf{Effect of Regularization Coefficient $\alpha$.}
As defined in Eq.~(6), the coefficient $\alpha$ mediates the trade-off between the behavioral cloning loss and Q-value regularization, thus modulating the contribution of the critic network during training. Figure~\ref{fig:score_vs_alpha} demonstrates that the value of $\alpha$ in the actor loss significantly impacts the final performance. Specifically, we observe that setting $\alpha=1.0$ yields the highest score, while both lower and higher values result in reduced performance. This finding underscores the importance of appropriately selecting $\alpha$, with $\alpha=1.0$ providing the optimal balance between policy regularization and expressiveness.

\textbf{Effect of the Number of RTG and Action Samples.}

Figure~\ref{fig:score_mn} shows how varying the number of RTG and action samples in the dual-exploration module affects final performance. Both RTG and Action curves achieve peak scores when the sample number is 3, indicating that moderate stochastic exploration ($M=N=3$) benefits policy performance. Increasing the number further leads to stable or slightly fluctuating results, possibly because excessive sampling causes actions to deviate too much. These results suggest that using 3 samples achieves a good trade-off between exploration and stability, thus optimizing overall performance.

\section{Online A/B Test}

To evaluate the practical effectiveness of \ProjectName{} in real-world advertising systems, we conducted an online A/B test on the Direct Express platform of Taobao, one of the largest e-commerce platforms in China. Experimental setup and results are presented below; more details are in Appendix~\ref{app:Deploy Details of Online A/B Test}.

\subsection{Experimental Setup}
The online experiment is divided into two phases. We compare the performance of \ProjectName{} (treatment group) with the baseline DT (control group). The specific setup is as follows:

\begin{itemize}[leftmargin=10pt]
    \item \textbf{Experiment Duration}: 
    \begin{itemize}
        \item Phase 1: Regular days
        \item Phase 2: Promotion period
    \end{itemize}

    \item \textbf{Metrics}: 
    \begin{itemize}
        \item \textit{Ad Click}: Number of clicks on ads
        \item \textit{Ad Cost}: Total advertising cost via bidding
        \item \textit{Ad GMV}: Gross Merchandise Volume from ad clicks
        \item \textit{Ad ROI}: Return on Investment
        \item \textit{Global Click}: Total platform clicks (organic + ads)
        \item \textit{Global GMV}:  GMV from non-ad (organic) clicks
    \end{itemize}

\eat{
    \item \textbf{Metrics}: 
    \begin{itemize}
        \item \textit{Ad Click}: Number of user clicks received by sponsored ads through the bidding system.
        \item \textit{Ad cost}: Total cost incurred by the bidding system.
        \item \textit{Ad GMV}: Gross Merchandise Volume directly attributed to ad clicks.
        \item \textit{Ad ROI}: Ratio of \textit{Ad GMV} to \textit{Ad cost}.
        \item \textit{Global Click}: The total number of clicks on the platform, including clicks from non-advertising channels such as organic search and recommendation.
        \item \textit{Global GMV}: GMV from organic search and recommendation traffic (non-ad clicks).
    \end{itemize}
} 
    \item \textbf{Control Strategy}: 
    The control group uses Original DT for different product categories, with both groups aligned on intervention strategies except for the bidding policy.
\end{itemize}

\subsection{Experimental Results}
\begin{figure}[!t]
    \centering
    % 第一个子图
    \begin{subfigure}[b]{\linewidth}
        \centering
        \includegraphics[width=\linewidth]{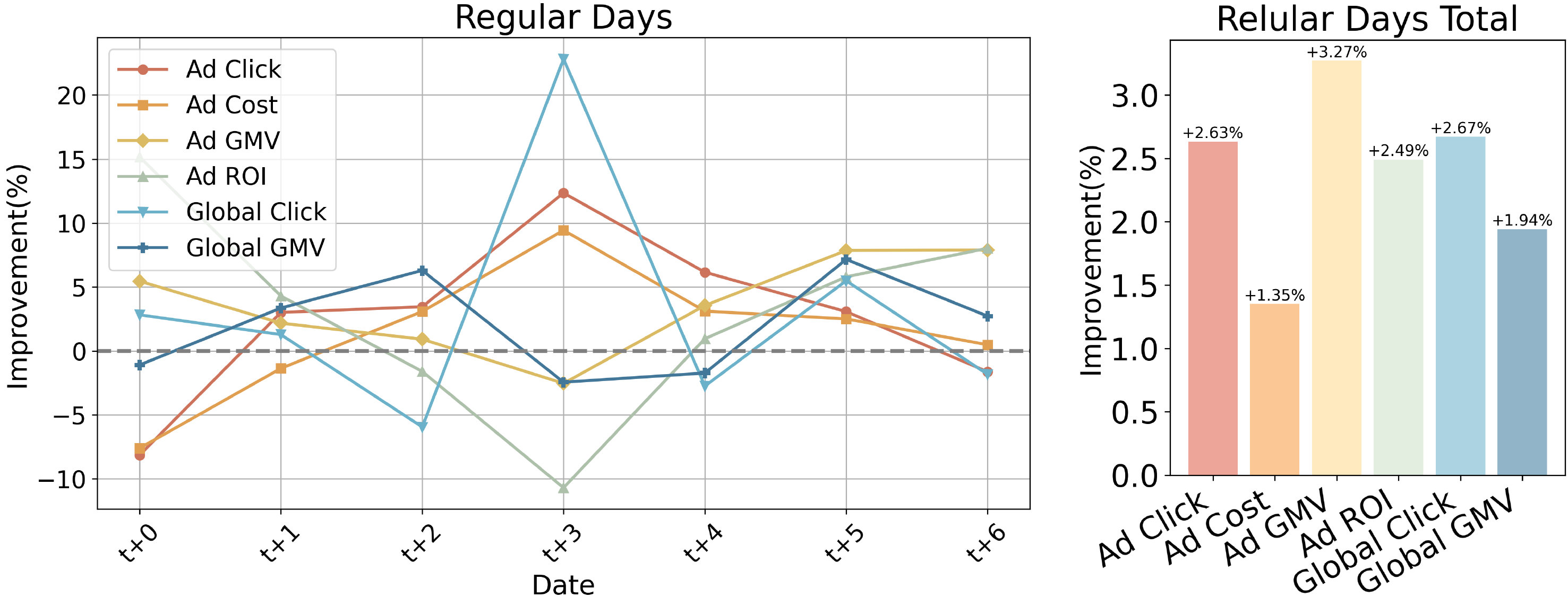}
        \caption{Online A/B Test in Regular Days}
        \label{fig:Regular}
    \end{subfigure}
    
    \vspace{8pt} % 可根据需要调整上下间隔

    % 第二个子图
    \begin{subfigure}[b]{\linewidth}
        \centering
        \includegraphics[width=\linewidth]{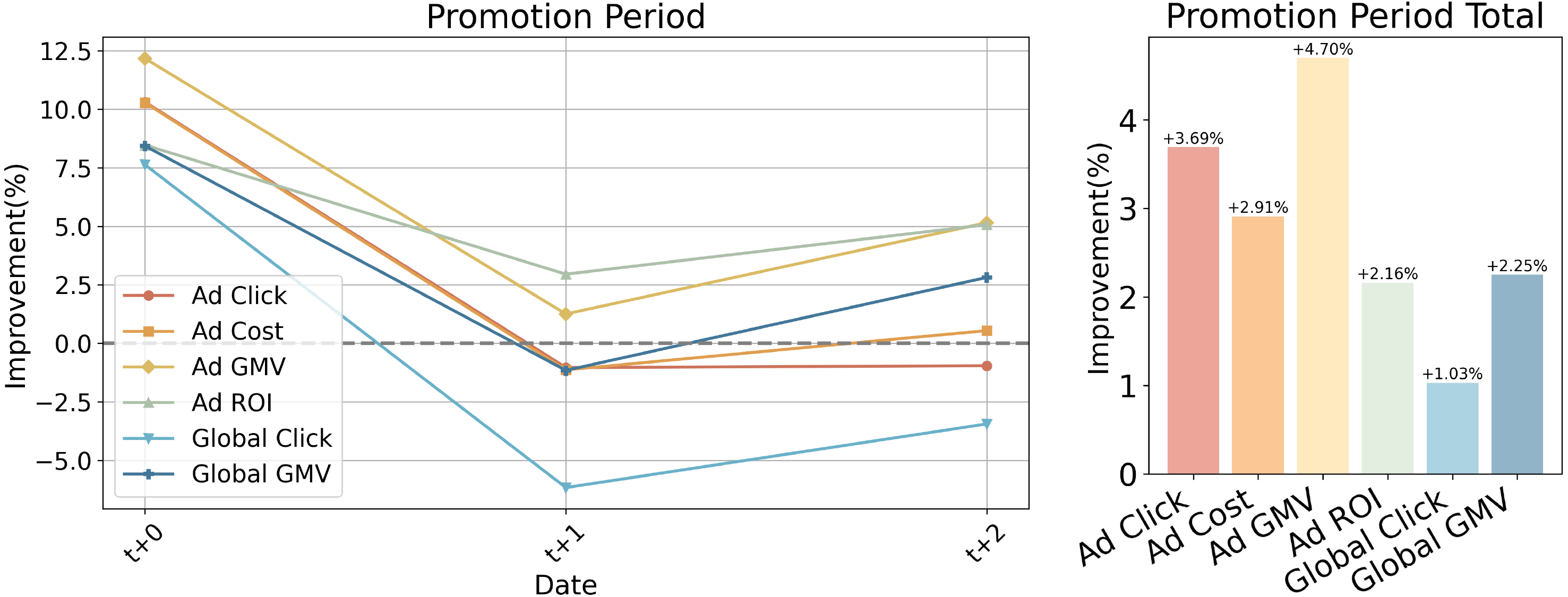}
        \caption{Online A/B Test in Promotion Period}
        \label{fig:promotion}
    \end{subfigure}
    \caption{Online A/B Test}
    \label{fig:AB}
\end{figure}

Figure~\ref{fig:AB} summarizes the key performance metrics for both phases. The following observations can be drawn: 1)~\textbf{Generalization Across Scenarios}. The method performs well during both the regular period (Phase 1) and the promotion period (Phase 2), demonstrating strong adaptability to fluctuations in traffic volume and changes in user behavior; 2)~\textbf{Cost Efficiency}. In Phase 1, the experimental group experiences a slight increase in Ad Cost (+1.35\%), while Ad GMV rises by 3.27\%, resulting in a 2.49\% improvement in Ad ROI. In Phase 2, although Ad Cost increases by 2.91\%, Ad GMV grows by 4.70\%, still achieving a 2.16\% uplift in ROI. This indicates flexible cost allocation under varying traffic conditions; 3)~\textbf{Exploration Benefits}. In Phase 2, with a significant increase in Ad Cost, both Ad GMV and ROI increase noticeably, indicating that the exploration mechanism effectively identifies previously under-exploited bidding strategies, especially in high-traffic scenarios.

The online A/B test confirms that \ProjectName{} significantly outperforms traditional DT method in real-world advertising systems. It achieves consistent improvements in core metrics while maintaining budget discipline, demonstrating its readiness for full-scale deployment. These results validate the effectiveness of integrating Q-value regularization and dual-exploration mechanism into generative decision models for complex auto-bidding tasks in advertising.

\section{Conclusion}
We propose a Q-value regularized generative auto-bidding method that augments the Decision Transformer with Q-regularization and dual-exploration mechanism. \ProjectName{} enables the learning of superior bidding policies from offline data and effective exploration of out-of-distribution policies. Extensive experiments on public datasets, simulation environments, and large-scale online A/B tests demonstrate significant improvements over prior methods across key business metrics. \ProjectName{} offers a practical and robust solution for auto-bidding in real-world advertising systems.

However, \ProjectName{} has following limitations. In complex environments involving multiple advertisers with unknown competitor bids, the exploration mechanism may not fully capture subtle environmental shifts. Additionally, the model lacks stability measures for abrupt changes in traffic conditions, such as those during promotions. Future work will explore fine-grained control strategies to enhance robustness in these scenarios.

\begin{acks}
This work was supported by Alibaba Group through Alibaba Innovative Research Program; This work was supported by National Natural Science Foundation of China (No. 62272349, No. 62302345, No. U23A20305);
\end{acks}

\bibliographystyle{ACM-Reference-Format}
\balance
\bibliography{sample-base}

\appendix
\section{More details about AuctionNet benchmark}
\subsection{Offline Dataset}
\label{app:Offline Dataset}
The AuctionNet dataset\footnote{https://github.com/alimama-tech/AuctionNet} \cite{su2024auctionnet} is collected from the logs of an advertising bidding environment, where multiple automated bidding agents compete for arriving impression opportunities. Each period is divided into 48 independent time steps, with approximately 480,000 impression opportunities per time step. The dataset consists of a dense subset and a sparse subset, which mainly differ in terms of conversion rates and CPA range. Notably, \textbf{the conversion rate of the sparse subset more closely reflects realistic scenarios}, whereas the conversion rate of the dense subset is approximately an order of magnitude higher than typically observed in practice. Key statistics of the dataset are summarized in Table~\ref{tab:data-statistics}.

\begin{table}[htbp]
    \centering
    \caption{Data statistics.}
    \label{tab:data-statistics}
    \begin{tabular}{l|c|c}
        \hline\hline
        Parameters & AuctionNet & AuctionNet-Sparse \\
        \hline
        Trajectories               & 479,376    & 479,376     \\
        Delivery Periods           & 9,987      & 9,987       \\
        Time steps in a trajectory & 48         & 48          \\
        State dimension            & 16         & 16          \\
        Action dimension           & 1          & 1           \\
        Return-To-Go Dimension     & 1          & 1           \\
        Budget range               & [450, 11850] & [2000, 4850] \\
        Action range               & [0, 493]   & [0, 589]    \\
        Impression's value range   & [0, 1]     & [0, 1]      \\
        CPA range                  & [6, 12]    & [60, 130]   \\
        Total conversion range     & [0, 1512]  & [0, 57]     \\
        Average conversion rate    & 0.0051           & 0.0005       \\
        \hline\hline
    \end{tabular}
\end{table}

In addition, for each subset, AuctionNet provides both impression-level training data and trajectory training data. The trajectory data is aggregated from the impression-level data, with the formats described as follows:

\textbf{Description of Impression-level Training Data Format}

\begin{itemize}
    \item deliveryPeriodIndex: Represents the index of the current delivery period.
    \item advertiserNumber: Represents the unique identifier of the advertiser.
    \item advertiserCategoryIndex: Represents the index of the advertiser's industry category.
    \item budget: Represents the advertiser's budget for a delivery period.
    \item CPAConstraint: Represents the CPA constraint of the advertiser.
    \item timeStepIndex: Represents the index of the current decision time step.
    \item remainingBudget: Represents the advertiser's remaining budget before the current step.
    \item pvIndex: Represents the index of the impression opportunity.
    \item pValue: Represents the conversion action probability when the advertisement is exposed to the customer.
    \item pValueSigma: Represents the prediction probability uncertainty.
    \item bid: Represents the advertiser's bid for the impression opportunity.
    \item xi: Represents the winning status of the advertiser for the impression opportunity, where 1 implies winning the opportunity and 0 suggests not winning the opportunity.
    \item adSlot: Represents the won ad slot. The value ranges from 1 to 3, with 0 indicating not winning the opportunity.
    \item cost: Represents the cost that the advertiser needs to pay if the ad is exposed to the customer.
    \item isExposed: Represents whether the ad in the slot was displayed to the customer, where 1 implies the ad is exposed and 0 suggests not exposed.
    \item conversionAction: Represents whether the conversion action has occurred, where 1 implies the occurrence of the conversion action and 0 suggests that it has not occurred.
    \item leastWinningCost: Represents the minimum cost to win the impression opportunity, i.e., the 4th highest bid of the impression opportunity.
    \item isEnd: Represents the completion status of the advertising period, where 1 implies either the final decision step of the delivery period or the advertiser's remaining budget falling below the system-set minimum remaining budget.
\end{itemize}

\textbf{Description of Trajectory Data Format}

\begin{itemize}
    \item \textbf{deliveryPeriodIndex}: Indicates the index of the current delivery period.
    \item \textbf{advertiserNumber}: The unique identifier for the advertiser.
    \item \textbf{advertiserCategoryIndex}: The industry category index of the advertiser.
    \item \textbf{budget}: The advertiser's budget for one delivery period.
    \item \textbf{CPAConstraint}: The advertiser's CPA (cost per action) constraint.
    \item \textbf{realAllCost}: The total cost incurred by the advertiser during the delivery period.
    \item \textbf{realAllConversion}: The total number of conversions generated by the advertiser during the delivery period.
    \item \textbf{timeStepIndex}: The index of the current decision step.
    \item \textbf{state}: The state at the current decision step.
    \item \textbf{action}: The action taken at the current decision step.
    \item \textbf{reward}: The sparse reward at the current decision step (i.e., the sum of conversions in the current step).
    \item \textbf{reward\_continuous}: The continuous reward at the current decision step (i.e., the sum of pValues for all impression traffic in the current step).
    \item \textbf{done}: Indicates whether the delivery period is completed; 1 denotes either the last step of the period or the remaining budget of the advertiser falls below the minimum threshold set by the system.
    \item \textbf{next\_state}: The state at the next decision step.
\end{itemize}

\subsection{Simulation Environment}
\label{app:Simulation Environment}
AuctionNet~\cite{su2024auctionnet} implements a standardized simulated online bidding environment. The bidding models to be evaluated are required to represent a specific advertiser and compete against baseline models provided by the environment, under given budget and CPA constraints. The simulation environment runs the bidding models multiple times with different advertiser configurations and various simulated traffic data. 

Specifically, the bidding model operates on 48 advertisers, and each advertiser is evaluated over 7 rounds. Each advertiser bids independently and has no knowledge of other advertisers’ bidding information. The simulated online bidding environment is a dynamic game environment, where the auction process is conducted in real time based on the bids from different models, whereas in the offline evaluation, the winner is determined only by the historical minimum winning price. This enables the simulation environment to provide more realistic testing than offline testing.

\section{Deploy Details of Online A/B Test}
\label{app:Deploy Details of Online A/B Test}

To evaluate the effectiveness of \ProjectName{}, we implemented it on Taobao within the Multiple Constraint Bidding (MCB) framework. In this setting, advertisers specify a budget and may optionally impose Cost Per Click (CPC) or Return on Investment (ROI) constraints. The bidding strategy aims to maximize traffic value while satisfying these requirements. For comparison, the control group adopted the original DT, with both groups employing identical intervention strategies except for the bidding policies. The experimental setup is outlined as follows:

\begin{itemize}[leftmargin=10pt]
    \item \textbf{Action}: To stabilize bidding outcomes, the bid coefficient $\lambda$ is determined using a windowed average over the preceding two hours, which comprises $E$ time steps, as follows:
    \[
        \lambda_{t} = a_{t} + \frac{1}{|E|} \sum_{t' = t-E}^{t-1} \lambda_{t'}
    \]
    where $a_{t}$ represents the action output by \ProjectName{} at time step $t$.
    
    \item \textbf{State}: A total of 19 state features are collected by recording the interactions between online ad bids and the platform's bidding environment:
    \begin{itemize}
        \item \textbf{Campaign status}: Proportion of remaining bidding steps, proportion of remaining budget, and deviation from CPC.
        \item \textbf{Historical and recent state averages (over the last three time steps)}: Number of impressions, number of clicks, number of conversions, advertising GMV (Gross Merchandise Value), click-through rate (CTR), conversion rate(CVR), ad cost, and bid.
    \end{itemize}
    
    \item \textbf{Return-to-Go}: The reward function is defined as the expected GMV from the current time step to the end of the day, under the constraints of CPC and budget.
\end{itemize}

For training, we collected data from the most recent seven days, yielding approximately 130{,}000 trajectories, each comprising 48 time steps. Online A/B testing was performed over a consecutive ten-day period, which included seven regular days and three promotional days. For each MCB campaign, 33\% of the product items were allocated to \ProjectName{}.The proposed method consistently demonstrates strong adaptability during both regular and promotional periods, achieving notable improvements in Ad GMV (up to 4.70\%) and ROI (up to 2.49\%) with only a slight increase in Ad Cost. These results suggest that the method enables effective and flexible cost allocation, as well as improved identification of optimal bidding strategies in dynamic traffic environments.

\end{document}